\documentclass[11pt]{article}

\usepackage[final]{acl}

\usepackage{times}
\usepackage{latexsym}

\usepackage[T1]{fontenc}

\usepackage[utf8]{inputenc}

\usepackage{microtype}

\usepackage{inconsolata}

\usepackage{listings}
\usepackage{xcolor}
\lstdefinestyle{prompt}{
    basicstyle=\footnotesize\ttfamily,
    breaklines=true,
    breakatwhitespace=true,
    frame=single,
    framesep=3pt,
    xleftmargin=4pt,
    xrightmargin=4pt,
    keepspaces=true,
    columns=flexible,
}

\usepackage{graphicx}
\graphicspath{{scripts/figures/}}

\usepackage{booktabs}

\usepackage{multirow}

\title{FinRAG-12B: A Production-Validated Recipe for Grounded Question Answering in Banking}

\author{
Denys Katerenchuk\textsuperscript{1},
Pablo Duboue\textsuperscript{2}\thanks{Work done while at Kasisto},
Keelan Evanini\textsuperscript{3}\footnotemark[1], 
David Gondek\footnotemark[1], \\
\textbf{Nithin Govindugari\textsuperscript{1},
Olivier Allauzen\textsuperscript{1},
Joshua Baptiste\textsuperscript{1},
David J More\textsuperscript{1},
Joshua Schechter\textsuperscript{1}} \\
\textbf{\textsuperscript{1}Backbase, New York, NY} \\
\textsuperscript{2}Textualization, Vancouver, Canada \\
\textsuperscript{3}NBME, Philadelphia, PA \\
\texttt{denys.katerenchuk@backbase.com, pablo.duboue@gmail.com, kevanini@nbme.org}
}

\begin{document}
\maketitle

\begin{abstract}
Large language models (LLMs) are rapidly being adopted across various domains.
However, their adoption in banking industry faces resistance due to demands
for high accuracy, regulatory compliance, and the need for verifiable and grounded responses.
We present a unified, data-efficient framework for training grounded domain-specific LLMs that 
optimizes answer quality, citation grounding, and calibrated refusal under real-world deployment constraints.
First, we describe a data generation pipeline that combines LLM-as-a-Judge
filtering, citation annotation, and curriculum learning with only 143M tokens.
The resulting 12B model achieves high answer quality outperforming GPT-4.1 on 
citation grounding, with a modest citation tradeoff versus the untuned base. 
Second, we propose a calibrated refusal mechanism: training on 22\%
unanswerable examples yield a 12\% ``I don't know'' rate, substantially improving over 
the base model’s unsafe 4.3\% rate while avoiding GPT-4.1’s over-refusal (20.2\%). 
Third, we present an end-to-end methodology spanning from data curation to quantized serving. 
The system is deployed at 40+ financial institutions, achieving a 7.1 percentage point improvement 
in query resolution ($p < 0.001$). Additionally, the model delivers 3–5x faster responses at 20–50x lower cost compared to GPT-4.1.
\end{abstract}

\section{Introduction}
Large language models (LLMs) have transformed natural language processing 
across a wide range of applications, including customer support, content generation, 
and code synthesis. However, their adoption in the regulated banking industry remains 
limited due to their tendency to hallucinate, exhibit over-agreeable behavior, and 
lack alignment with domain-specific knowledge and constraints. 

In addition, banking applications operate over dynamic, frequently changing data, 
such as interest rates, account balances, and institutional policies, requiring 
models to ground their responses in up-to-date information retrieved from internal 
or external sources. This setting places strict demands on factual accuracy, traceability, and latency.

Consider a customer asking a bank’s virtual assistant about early mortgage payoff penalties. 
The system must retrieve the latest policy, interpret it correctly, and generate a response 
grounded in source documents within seconds. If the relevant information is unavailable, the model must explicitly refrain from answering rather than fabricate an answer, as incorrect responses may lead to regulatory and financial risk. Achieving this level of reliability and responsiveness requires moving beyond off-the-shelf LLMs toward domain-specialized, grounded systems.

To address these challenges, we introduce FinRAG-12B, a 12B-parameter LLM optimized for retrieval-augmented generation (RAG) in banking. We propose a unified, data-efficient training framework that jointly improves answer quality, citation grounding, and calibrated refusal. Our approach centers on a multi-stage data curation pipeline combining LLM-as-a-Judge filtering, citation annotation, and curriculum learning, achieving 73\% citation quality with only 143M tokens.

Through systematic ablations, we show that incorporating 22\% unanswerable examples enables calibrated “I don’t know” responses, substantially reducing hallucinations while avoiding excessive refusal on answerable queries. Finally, we present a complete, production-ready methodology from data curation to quantized serving that was validated through deployment across 40+ financial institutions.
Our contributions are threefold:
(1) a high-quality, data-efficient pipeline for generating grounded training data;
(2) a calibrated refusal strategy based on controlled negative sampling to mitigate hallucinations; and
(3) an end-to-end framework for training and deploying LLMs in production RAG systems for banking.

\section{Related Work}

The field of financial NLP has advanced rapidly in recent years. BloombergGPT \cite{wu2023bloomberggpt} trained a 50B-parameter model on 346B tokens of proprietary financial data, achieving strong performance on tasks such as sentiment analysis, named entity recognition, and question answering. However, the model was not publicly released, and the work does not report deployment-critical metrics such as latency, inference cost, or hallucination rates. FinGPT \cite{yang2023fingpt} demonstrated that parameter-efficient fine-tuning (e.g., LoRA) can achieve competitive results on financial sentiment analysis at low cost, but focuses primarily on classification tasks rather than generative RAG with citation requirements. Earlier work such as FinBERT \cite{araci2019finbert} established the importance of domain adaptation, showing that continued pretraining on financial corpora improves downstream performance.

Retrieval-augmented generation (RAG) has emerged as a standard approach for improving factuality by grounding model outputs in external knowledge sources \cite{lewis2020retrieval}. While RAG reduces hallucinations, it introduces new failure modes: models may ignore retrieved context, rely on irrelevant passages, or generate unsupported claims. \citet{liu2024lost} further show that LLMs exhibit positional bias in long contexts, attending disproportionately to tokens at the beginning and end of the input. We mitigate this issue through randomized context placement during training, sampled from a mixture of discrete right trapezoidal harmonic decay distribution.

A growing body of work emphasizes the importance of data quality over scale in LLM training. Phi-3 \cite{abdin2024phi3} demonstrates that smaller models trained on carefully filtered data can match or exceed the performance of larger models trained on noisier corpora. Similarly, \citet{zhou2024lima} show that a small number of high-quality instruction examples can suffice for alignment. Subsequent work explores principled data selection strategies, including pairwise quality ranking \cite{wettig2024qurating}, loss-based example selection \cite{xia2024less}, and optimal data mixing \cite{ye2024datamixing}. These findings motivate our multi-stage training pipeline, which sequences data of varying quality and provenance rather than mixing all sources uniformly.

For parameter-efficient training, we build on LoRA \cite{hu2022lora} and DoRA \cite{liu2024dora}, which learn low-rank weight updates to adapt large models without full fine-tuning. DoRA extends LoRA via directional decomposition, improving training stability while reducing memory overhead. Prior work has also shown that curated datasets can outperform larger unfiltered corpora \cite{gunasekar2023textbooks}; we operationalize this insight through LLM-based quality filtering during data curation.

Curriculum learning has been shown to improve generalization by presenting training examples in order of increasing difficulty \cite{bengio2009curriculum}. In domain adaptation settings, \citet{gururangan2020dont} demonstrate that continued pretraining on domain-specific corpora improves downstream task performance. We adopt a two-stage curriculum: Stage~1 adapts the model to general financial language using open-source data, while Stage~2 specializes the model on proprietary banking interactions. This design also enhances reproducibility, as Stage~1 can be replicated using publicly available datasets. Unlike prior work, we integrate data curation, grounding, and refusal calibration into a unified framework optimized for real-world deployment in regulated environments.

\section{System Architecture}

\subsection{Base Model Selection}

The choice of base model is critical for balancing performance and deployment constraints. We require strong instruction-following capabilities, with low hallucination rates alongside efficiency in latency and cost. After surveying models such as Phi-4 14B and Qwen3 14B, we select Gemma 3 12B-IT \cite{team2024gemma} as it provides competitive instruction-following performance, a 128K context window, and permissive commercial licensing while remaining computationally efficient.

\subsection{Training Data Pipeline}

Training data quality is a primary determinant of LLM performance, particularly in high-stakes domains such as banking. While financial institutions possess large volumes of data, much of it is noisy, outdated, or contains personally identifiable information (PII), limiting its usability.

To address this, we design a multi-stage data curation pipeline that prioritizes quality over scale. The final corpus consists of 98,648 samples (143M tokens) drawn from a combination of open-source and proprietary data, filtered and processed through multiple quality control stages. The entire training data set is shown in Table \ref{tab:data}. Each dataset required postprocessing to make sure that we align data with our requirements.

\begin{table}[h]
\centering
\small
\begin{tabular}{lrr}
\toprule
\textbf{Source} & \textbf{Samples} & \textbf{License} \\
\midrule
RAG-v1 (Open) & 43,581 & Apache 2.0 \\
SEC Reports (Synthetic QA) & 16,773 & Public \\
CommonCrawl (Financial) & 20,499 & CC0 \\
Refusal Calibration (Proprietary) & 17,795 & Internal \\
\midrule
\textbf{Total} & \textbf{98,648} & -- \\
\bottomrule
\end{tabular}
\caption{Training data composition. Stage 1 uses open-source data (RAG-v1,
CommonCrawl). Stage 2 adds proprietary banking conversations and synthetic
QA from SEC filings.}
\label{tab:data}
\end{table}

\subsubsection{RAG-v1} 
We curate 43,581 samples from the \texttt{glaiveai/RAG-v1} dataset,\footnote{https://huggingface.co/datasets/glaiveai/RAG-v1} a synthetic dataset designed for retrieval-augmented generation. Each sample includes a question, supporting documents, and a cited answer. We apply JudgeLM \cite{zhu2023judgelm} to filter out low-quality responses (score $<5$), ensuring consistent supervision quality.

\subsubsection{Synthetic SEC QA}
\label{sec:sec-synth-qa}
Purely synthetic QA generation pipelines often suffer from distribution mismatch with real user queries, regurgitative artifacts, and limited control over difficulty and grounding \cite{zhang2024regurgitative}. To address these limitations, we propose a style-conditioned, multi-stage QA generation pipeline grounded in financial SEC filings (10-K and 10-Q), producing 16,773 high-quality training samples.

To align synthetic data with real-world conditions, we design a five-stage generation pipeline:
\begin{enumerate}
\item We segment each SEC filing (10-K, 10-Q) into passage-length chunks (400–600 tokens) using semantic boundaries when available and fixed windows otherwise. This ensures that each example reflects the granularity of retrieved context in real RAG systems while preserving local coherence.
\item For each passage, we generate questions at four difficulty levels (easy, medium, hard, expert) via prompt conditioning. Difficulty is operationalized through required reasoning depth: easy questions target explicit facts, medium require light synthesis, hard involve multi-sentence reasoning, and expert questions require implicit inference or domain knowledge. We observe that  easy and medium difficulty questions most closely match real user queries and therefore assign them the highest sampling weight during training.
\item To reduce distribution mismatch, we rephrase generated questions using few-shot prompting conditioned on empirical query distributions. Specifically, we control for (i) query style (e.g., fragment, “how-do-I”, “what-is”), (ii) length (sampled from a log-normal distribution), and (iii) formality. This transforms verbose, interrogative LLM outputs into concise, user-like queries (e.g., converting “What is the minimum credit score required for mortgage approval?” $\rightarrow$ “min credit score for mortgage”).
\item Answers are generated using the original question and its corresponding gold passage, with explicit instructions to ground all claims in the provided context. We enforce citation-style outputs during generation, ensuring that each answer is traceable to specific source spans. This step establishes supervision for grounded generation rather than parametric recall.
\item To simulate realistic retrieval noise, we inject 3–7 distractor passages sampled from topically similar documents. The gold passage is randomly interleaved among distractors, with positions drawn from a mixture of discrete right trapezoidal harmonic decay distribution. This exposes the model to varying context positions and trains it to identify and ground responses in the correct evidence rather than relying on positional heuristics.
\end{enumerate}

Compared to single-shot generation, this pipeline substantially improves alignment with real query distributions, reducing question-type divergence by 10× (JS: 0.434 → 0.041) and matching average query length (8.85 vs. 9.91 words). It also increases lexical diversity and domain coverage (Table~\ref{tab:qgen-key}). These results highlight the importance of jointly modeling grounding and query distribution for effective RAG training.

\begin{table}[htbp]
\centering
\small
\begin{tabular}{@{}lccc@{}}
\toprule
\textbf{Metric} & \textbf{Single} & \textbf{Multi} & \textbf{Real} \\
\midrule
Avg.\ length (words)              & 19.55 & \textbf{8.85}  & 9.91  \\
Jaccard w/ real ($\uparrow$)      & 0.098 & \textbf{0.140} & —     \\
Type entropy ($\uparrow$)         & 1.418 & \textbf{1.745} & 2.281 \\
Type JS div.\ ($\downarrow$)      & 0.434 & \textbf{0.041} & —     \\
Coverage, cos.\ ($\uparrow$)      & 0.464 & \textbf{0.520} & —     \\
Distinct-2 ($\uparrow$)           & 0.295 & \textbf{0.451} & 0.721 \\
Fin.\ term recall ($\uparrow$)    & 0.902 & \textbf{0.951} & —     \\
\bottomrule
\end{tabular}
\caption{Single-shot vs.\ multi-step QA generation pipeline evaluated on lexical, question-type, semantic, diversity, and domain metrics. $\uparrow$\,=\,higher is better, $\downarrow$\,=\,lower is better.}
\label{tab:qgen-key}
\end{table}

\subsubsection{CommonCrawl Financial Subset} 
Because user data is subject to strict privacy constraints, we adopt a hybrid approach
to construct the training set without exposing personally identifiable information.
First, we train a random forest classifier to identify banking-relevant content.
Next, we extract real user questions that contain no PII and appear more than $n$ times,
preventing memorization of rare, potentially sensitive queries.
Finally, we cross-reference these questions with the classifier to retrieve relevant passages.
The resulting dataset of 20,499 samples is grounded in real-world usage while remaining
compliant with privacy requirements.

To incorporate real-world signal without exposing sensitive data, we construct a privacy-compliant dataset from CommonCrawl. We train a classifier to identify banking-relevant content, extract frequently occurring non-PII user queries, and retrieve aligned passages.

We apply a similar multi-stage processing strategy as in Section~\ref{sec:sec-synth-qa}, including context grounding, to ensure consistency with the synthetic QA pipeline. This yields 20,499 samples that capture real-world usage patterns while maintaining strict privacy constraints.

\subsubsection{Banking Refusal Calibration Data}
In production RAG systems, retrieved context may be incomplete or insufficient to answer a user’s query. In such cases, models must abstain rather than generate unsupported responses. However, instruction-tuned LLMs exhibit a well-documented tendency toward over-compliance (sycophancy) \cite{sharma2024towards}, often producing plausible but ungrounded answers even in the absence of supporting evidence, thereby increasing hallucination risk \cite{lewis2020retrieval}. In regulated domains such as banking, such behavior can lead to compliance violations and undermine user trust.

To address this, we construct a dedicated refusal calibration dataset derived from real banking conversations. Each example pairs a user query with retrieved context that is topically relevant but lacks sufficient information to support a correct answer. The target output in these cases is an explicit “I don’t know” response, providing supervision for abstention behavior.

This subset comprises 17,795 examples (22\% of the training mixture), enabling controlled calibration of refusal behavior. By incorporating these negative examples during fine-tuning, we shift the model toward conservative, evidence-based responses: the model learns to ground its outputs in retrieved context when possible and to abstain when that context is insufficient, rather than relying on potentially outdated or incorrect parametric knowledge.

\subsection{Positional Bias}
RAG models exhibit positional bias, disproportionately attending to earlier or later context segments \cite{liu2024lost}. To mitigate this, we randomize the placement of relevant passages among distractors during training. Specifically, we sample positions from a mixture of discrete right trapezoidal harmonic decay distribution defined as:
$$P(X = x) = \frac{1}{N - K_{\min} + 1} \sum_{K = \max(x,\, K_{\min})}^{N} \frac{1}{K}$$
This strategy reduces systematic bias toward fixed context positions and encourages more uniform attention over retrieved evidence.

\subsection{Model Training}


We adopt a two-stage curriculum inspired by domain-adaptive pretraining \cite{gururangan2020dont}, progressively transitioning from grounded domain adaptation to task-specific specialization (Table~\ref{tab:curriculum}).

In Stage~1 (Domain Adaptation), the model is trained on 60,354 samples from RAG-v1 and SEC-derived synthetic QA data using a conservative learning rate of $1 \times 10^{-6}$ with cosine decay. This stage establishes foundational capabilities for grounded generation, including citation alignment, evidence-based reasoning, and familiarity with financial text, using largely reproducible data sources.

In Stage~2 (Task Specialization), training continues from the Stage~1 checkpoint on 38,294 samples drawn from CommonCrawl financial data and proprietary banking conversations at a higher learning rate of $5 \times 10^{-6}$ with linear decay. This stage introduces real-world query distributions and production-specific behaviors, including calibrated refusal, robustness to noisy retrieval, and institution-specific formatting.

We find that staged training is critical: jointly training on all data leads to degraded performance (Section~\ref{sec:curriculum-ablation}), underscoring the importance of separating grounded domain learning from real-world adaptation.

\begin{table}[h]
\centering
\small
\begin{tabular}{@{}llrr@{}}
\toprule
\textbf{Stage} & \textbf{Data Source} & \textbf{Samples} & \textbf{LR} \\
\midrule
1 & RAG-v1 + SEC Synth. QA & 60,354 & $1 \times 10^{-6}$ \\
2 & CC + Propriet. Banking & 38,294 & $5 \times 10^{-6}$ \\
\bottomrule
\end{tabular}
\caption{Curriculum learning stages. Stage~1 uses grounded, reproducible data for domain adaptation; Stage~2 introduces real-world and proprietary data for task specialization.}
\label{tab:curriculum}
\end{table}

\subsection{Refusal Calibration}

Calibrating a model to abstain when retrieved context is insufficient requires careful control of the negative-example ratio. An insufficient proportion of negative samples causes the model to default to its sycophantic prior \cite{sharma2024towards}, producing plausible but unsupported responses (high false-positive rate). Conversely, an excessive proportion leads to over-refusal, where the model abstains even when sufficient evidence is available (high false-negative rate).
To identify the optimal balance, we sweep the negative ratio from 10\% to 30\% in 2-percentage-point increments. We find that a ratio of 22\% yields the best trade-off, minimizing unsupported answers while preserving recall on answerable queries. Beyond 26\%, performance degrades due to excessive conservatism, with a sharp decline in recall.

\subsection{Training Configuration}

We fine-tune the model using LoRA \cite{hu2022lora} applied to all attention and MLP layers ($r{=}64$, $\alpha{=}256$, dropout$,{=},0.05$). Optimization is performed with 8-bit AdamW using a learning rate of $2 \times 10^{-5}$. We use a per-device batch size of 4 with gradient accumulation over 4 steps, resulting in an effective batch size of 16, and a maximum sequence length of 16,384 tokens. We reserve 1\% of the training data as a validation set and apply early stopping with a patience of 5 evaluation steps, restoring the best-performing checkpoint based on validation loss.
Training completes in 1,400 steps, requiring approximately 360 GPU-hours on 8$\times$ RTX A6000 GPUs with a total training cost \$1,800 (360 GPU-hours at \$5/hour). 

\subsection{Quantization}

To meet production latency and memory constraints, we quantize the fine-tuned model to W4A16 (4-bit weights, 16-bit activations) using SmoothQuant \cite{xiao2023smoothquant}. This reduces the model footprint from 24,GB to 8.4,GB ($2.86{\times}$ compression), enabling efficient single-GPU deployment.
Despite aggressive compression, performance degradation is minimal: the quantized model retains over 99\% of full-precision citation quality, demonstrating that grounded generation remains robust under low-bit quantization.

\section{Evaluation}

\subsection{Dataset}

We evaluate on a proprietary banking test set and complement it with a public financial benchmark to assess generalization.

\paragraph{Proprietary Banking Test Set.}
We construct a dataset of 258 RAG examples from three financial institutions spanning retail (e.g., account inquiries, loan information, branch hours) and commercial banking (e.g., treasury services, merchant processing). To reflect real-world retrieval conditions, some of the examples are unanswerable, where the provided context is topically related but does not contain sufficient information, making “I don’t know” the correct response.
Each example consists of a user query, five retrieved passages (1–2 relevant and 3–4 distractors), human-annotated reference answers with citation labels, and a binary answerability label. This design enables joint evaluation of grounding quality and refusal behavior under realistic RAG settings.

\paragraph{Public Benchmark.}
To assess generalization beyond proprietary data, we additionally evaluate on FinanceBench \cite{islam2023financebench}, which consists of 150 questions over SEC filings requiring free-form, grounded responses. FinanceBench evaluates both factual retrieval and citation grounding, providing a standardized benchmark for comparison.

\subsection{Metrics}

We evaluate along three dimensions: answer quality, refusal calibration,
and latency.

JudgeLM \cite{zhu2023judgelm} scores each response on a 1--10 scale using an LLM judge
that assesses correctness, completeness, and coherence given the retrieved
sources. Citation Quality \cite{es-etal-2024-ragas} (0--100) is a composite of
faithfulness, source relevance, information synthesis, and source usage
(evaluation prompts in Appendix~\ref{sec:prompts-eval}).

We report QA F1, which combines precision (fraction of generated
answers that are correct) and recall (fraction of answerable queries that
receive an answer). F1 captures the tension between hallucination risk
(low precision) and over-caution (low recall). We additionally report the
refusal rate and the fraction of abstentions that are true negatives.

\section{Results}
\label{sec:results}
We compare FinRAG-12B against Gemma 3 12B-IT (no fine-tuning), GPT-4.1 (API). 
BloombergGPT weights are unavailable. FinGPT has no citation
evaluation.

\subsection{Answer Quality and Citation Grounding}

\begin{table}[h]
\centering
\small
\begin{tabular}{lcccc}
\toprule
\textbf{Model} & \textbf{Jud.LM} & \textbf{Cit.\ Q} & \textbf{QA F1} & \textbf{IDK\%} \\
\midrule
Gemma 3 12B & 5.70 & \textbf{80.2} & \textbf{0.964} & 4.3 \\
GPT-4.1 & 5.72 & 70.8 & 0.900 & 20.2 \\
\midrule
\textbf{FinRAG-12B} & \textbf{6.21} & 73.1 & 0.936 & \textbf{12.0} \\
\bottomrule
\end{tabular}
\caption{Main results on 258 banking QA examples (3 institutions).
JudgeLM: answer quality (1--10); Cit.\ Q: citation quality (0--100);
QA F1: precision--recall on answerable queries; IDK\%: abstention rate.}
\label{tab:main}
\end{table}

FinRAG-12B achieves the highest overall answer quality, with a JudgeLM score of 6.21, outperforming both GPT-4.1 (+0.49) and the base Gemma~3 model (+0.51). It also improves citation grounding relative to GPT-4.1, achieving a +2.3 point gain in citation quality (73.1 vs.\ 70.8).
While the base model reports a higher citation quality score (80.2), this is driven by a broad citation strategy that references multiple retrieved passages regardless of relevance. In contrast, FinRAG-12B is trained to cite selectively, grounding each claim in only the most relevant evidence.
Importantly, FinRAG-12B achieves a more balanced refusal behavior, with an abstention rate of 12.0\%, compared to under-refusal in the base model (4.3\%) and over-refusal in GPT-4.1 (20.2\%). This demonstrates effective calibration between answerability and abstention.

\subsection{Public Benchmark Results}

\begin{table}[h]
\centering
\small
\begin{tabular}{@{}lc@{}}
\toprule
\textbf{Model} & \textbf{FinanceBench F1} \\
\midrule
Gemma 3 12B (base) & 0.249 \\
GPT-4.1 (API) & 0.238 \\
\midrule
\textbf{FinRAG-12B} & \textbf{0.284} \\
\bottomrule
\end{tabular}
\caption{Results on FinanceBench (150 SEC filing questions).
FinRAG-12B achieves the highest F1 with a 97.3\% citation rate.}
\label{tab:public-benchmarks}
\end{table}

On FinanceBench \cite{islam2023financebench}, FinRAG-12B generalizes beyond proprietary data, achieving the highest F1 score (0.284) with a 97.3\% citation rate, outperforming both the base model (0.249) and GPT-4.1 (0.238).

\subsection{Curriculum Learning Ablation}
\label{sec:curriculum-ablation}

\begin{table}[h]
\centering
\small
\setlength{\tabcolsep}{4pt}
\begin{tabular}{@{}lccccc@{}}
\toprule
\textbf{Data Strategy} & \textbf{Jud.} & \textbf{QA F1} & \textbf{Cit.Q} & \textbf{IDK\%} & \textbf{TN\%} \\
\midrule
External only & 5.72 & \textbf{.972} & 76.1 & 0.4 & 0 \\
Internal only & 5.62 & .913 & 69.2 & 17.4 & 53 \\
Combined (all) & 3.28 & .706 & 51.2 & 46.5 & 39 \\
Curriculum (staged) & \textbf{5.91} & .938 & \textbf{74.7} & \textbf{13.2} & \textbf{56} \\
\bottomrule
\end{tabular}
\caption{Data strategy ablation on 258 banking QA examples.
Jud.: JudgeLM answer quality (1--10); Cit.Q: citation quality (0--100);
IDK\%: \% of queries refused; TN\%: \% of refusals that are correct.}
\label{tab:curriculum-ablation}
\end{table}

Table~\ref{tab:curriculum-ablation} isolates the contribution of each
data source. Training on external data alone yields high QA~F1 (0.972)
but the model almost never refuses (0.4\% IDK), making it unsafe for
unanswerable queries. Internal data alone teaches refusal (17.4\% IDK)
but citation quality drops to 69.2. Mixing all data simultaneously
collapses performance: JudgeLM falls to 3.28 and the model over-refuses
at 46.5\%, with only 39\% of refusals correct. The two-stage curriculum
resolves this conflict---external data first establishes citation
conventions, then internal data calibrates refusal---achieving the best
answer quality (5.91) and the highest true-negative precision (56\%)
among fine-tuned variants.

\subsection{Latency and Cost}

\begin{figure}[h]
\centering
\includegraphics[width=\columnwidth]{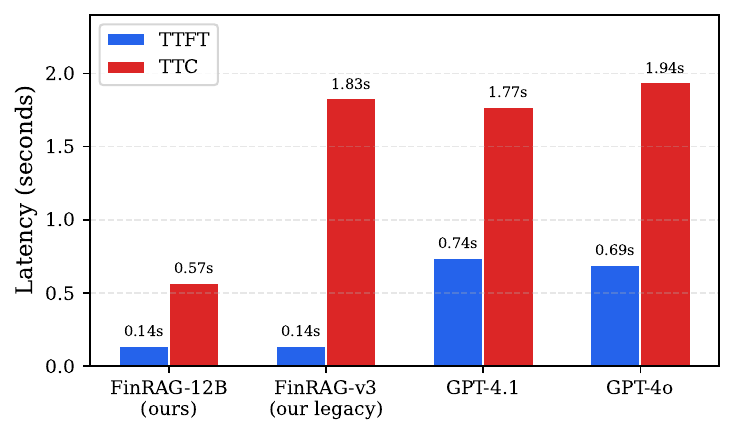}
\caption{Inference latency comparison. FinRAG-12B achieves 0.14s TTFT and 0.57s TTC, outperforming both proprietary APIs and the prior production model.}
\label{fig:latency}
\end{figure}

Figure~\ref{fig:latency} compares time to first token (TTFT) and total time to completion (TTC) in a RAG setting on a single RTX 6000 Ada GPU. FinRAG-12B achieves substantial latency improvements, operating 3--5$\times$ faster than GPT-4.1 across both metrics. Compared to the prior production model (FinRAG-v3, based on Mistral-7B-Instruct \cite{jiang2023mistral7b}), FinRAG-12B reduces TTC by 3.2$\times$ while maintaining comparable TTFT.
In addition to latency gains, FinRAG-12B offers significant cost advantages. Running on a single RTX 6000 Ada GPU, inference costs approximately \$0.001 per query at a throughput of 10K queries per day. In contrast, GPT-4.1 API pricing ranges from 
\$0.02 to \$0.05 per query, making it 20--50$\times$ more expensive.



\subsection{Production Impact}

While the preceding sections evaluate performance on curated benchmarks, production deployment provides the ultimate test of whether improvements translate into real-world outcomes. We analyze 3,297 randomly sampled user queries collected over seven months (May--December 2025) from a large U.S. credit union serving millions of retail banking customers, comparing a legacy production model against FinRAG-12B. Both models operate within the same RAG pipeline and share an identical knowledge base, ensuring that observed differences are attributable to the LLM rather than retrieval or data changes. 

\begin{table}[h]
\centering
\small
\setlength{\tabcolsep}{4pt}
\begin{tabular}{@{}lccc@{}}
\toprule
& \textbf{Old} & \textbf{FinRAG} & \textbf{$\Delta$} \\
\midrule
$N$ (queries) & 1,044 & 2,253 & --- \\
Resolution & 77.4\% & \textbf{84.5\%} & +7.1$^{***}$ \\
Unresolved & 20.7\% & \textbf{13.7\%} & $-$7.0$^{***}$ \\
\midrule
Satisfaction & 59.5\% & \textbf{62.9\%} & +3.4 \\
~~resolved & 65.0\% & \textbf{66.7\% }& +1.7 \\
~~unresolved & 23.7\% & \textbf{25.9\%} & +2.1 \\
\bottomrule
\end{tabular}
\caption{Production metrics (3,297 queries, 7 months).
Deltas in pp. $^{***}\!p{<}0.001$ ($\chi^2$).
Satisfaction: $p{=}0.26$, 95\% CI [$-$2.5, +9.3]\,pp.}
\label{tab:production}
\end{table}

As shown in Table~\ref{tab:production}, FinRAG-12B improves the resolution rate by 7.1 percentage points ($\chi^2{=}24.4$, $p{<}0.001$, Cramer’s $V{=}0.09$). This corresponds to 7 additional queries per 100 being resolved without human escalation. The unresolved rate decreases by a similar margin ($-$7.0 pp), indicating that the model converts previously unresolved queries into successful outcomes rather than redistributing errors across categories.

Overall user satisfaction increases from 59.5\% to 62.9\%, although this difference is not statistically significant ($\chi^2{=}1.3$, $p{=}0.26$; 95\% bootstrap CI [$-$2.5, $+$9.3] pp). To better understand this effect, we decompose satisfaction by outcome. Satisfaction remains nearly unchanged for both resolved queries (65.0\% vs.\ 66.7\%) and unresolved queries (23.7\% vs.\ 25.9\%), suggesting that per-response quality is largely comparable across models.
Instead, the overall improvement is driven by a distributional shift: a larger proportion of queries are resolved, moving users into a higher-satisfaction regime. This indicates that, in production RAG systems, increasing resolution rate rather than marginal improvements in response quality is the primary driver of user satisfaction.

\section{Conclusion}
We present FinRAG-12B, a citation-grounded LLM for banking RAG systems. We show that a data-efficient training pipeline achieves the highest answer quality (JudgeLM 6.21) while improving citation grounding over GPT-4.1 by 2.3 points.
We further demonstrate that incorporating 22\% unanswerable examples enables calibrated refusal behavior: FinRAG-12B abstains on 12\% of queries, balancing the under-refusal of the base model (4.3\%) and the over-refusal of GPT-4.1 (20.2\%).
These improvements translate directly to production. FinRAG-12B increases query resolution by 7.1 percentage points ($p{<}0.001$), while achieving 3--5$\times$ lower latency and 20--50$\times$ lower cost than commercial APIs. Notably, gains in user satisfaction are driven by improved resolution rates rather than per-response quality, highlighting resolution as the primary driver of user experience in RAG systems.
Our results show that data quality, grounding, and training methodology are key to building reliable LLMs for regulated domains.

\section*{Limitations}
Our evaluation focuses on banking RAG across three financial institutions (258 examples), and results may not generalize to other financial domains such as trading, insurance, or investment advisory. Additionally, the test set is skewed toward common retail banking queries, leaving rare edge cases underexplored.

The proprietary data used in Stage~2 training cannot be released due to privacy and regulatory constraints. However, Stage~1 relies entirely on open-source dataset (RAG-v1), and we can provide all training configurations. This enables researchers to reproduce our curriculum and adapt it to their own proprietary data.

Finally, our refusal evaluation captures explicit “I don’t know” responses and close variants, but may not fully account for hedged or partial uncertainty expressions (e.g., indirect or softened refusals), potentially underestimating nuanced abstention behavior.

\section*{Acknowledgements}
The authors thank David Gondek, Nithin Govindugari, Olivier Allauzen, Joshua Baptiste, DJ More, Binod Gyawali, Bob Stewart, George Trad, Jon Phillips, Priti Khanna, Victoria Povolotsky, Jared Kim, Samar Batra,  Romain Lebolloch, Gerard Benavides, Martin Lopez, Allison Ray, Yeonju Lee-Sikka, Max Schwartz, Tino Galizio, Rob Kassel, Jon Hopewell, Justin Arnoldi, Byron Wolff, Faisal Nematt, Robert Dugdale, Ben Ortega, Anand Venkatachalam, Mehul Pipalia, Raju Singh, Biswajit Singha, Michael Wasserfuhr, Helen Rosen and everyone on the team for their contributions to FinRAG-12B and the broader project at Kasisto.

We also thank the anonymous ACL reviewers and the meta-reviewer for their valuable feedback, which helped improve the quality of this manuscript.

This work was conducted under the executive leadership of Zor Gorelov, Sasha Caskey, Joshua Schechter, and Lance Berks. We are grateful for their support, trust, and vision.

\section*{Ethics Statement}

This work targets deployment in regulated financial services, where accuracy and reliability are critical. All proprietary training data was anonymized to remove personally identifiable information (PII) prior to use.
To mitigate the risks of hallucination, the model is explicitly trained to abstain when sufficient evidence is not available, aligning with regulatory requirements for verifiable responses.
As with all LLMs, there is a risk of inheriting biases present in training data. We monitor model behavior across user segments and continuously evaluate response quality to identify and mitigate potential disparities.

\bibliography{custom}

\appendix

\section{Training Hyperparameters}
\label{sec:hyperparams}

\begin{table}[h]
\centering
\small
\begin{tabular}{ll}
\toprule
\textbf{Parameter} & \textbf{Value} \\
\midrule
Base Model & Gemma 3 12B-IT \\
LoRA Rank ($r$) & 64 \\
LoRA Alpha ($\alpha$) & 256 \\
LoRA Dropout & 0.05 \\
Target Modules & q,k,v,o,gate,up,down \\
Learning Rate & $2 \times 10^{-5}$ \\
Optimizer & AdamW-8bit \\
Batch Size & 4 \\
Gradient Accumulation & 4 \\
Effective Batch Size & 16 \\
Max Sequence Length & 65,536 \\
Training Steps & 1,402 \\
Final Loss & 1.871 \\
\bottomrule
\end{tabular}
\caption{Training hyperparameters.}
\label{tab:hyperparams}
\end{table}

\section{Quantization Details}
\label{sec:quantization}

W4A16-G128 quantization configuration:

\begin{itemize}
    \item \textbf{Weight Quantization}: 4-bit, group size 128
    \item \textbf{Activation Precision}: 16-bit (FP16)
    \item \textbf{Preprocessing}: SmoothQuant, migration factor 0.5
    \item \textbf{Calibration}: 512 samples from training distribution
\end{itemize}

Size reduction: 24GB $\rightarrow$ 8.4GB (2.86$\times$ compression).
Citation quality degrades marginally, retaining over 99\% of full-precision performance.

\section{Data Generation Prompts}
\label{sec:prompts-datagen}

All synthetic data is generated by leveraging open weight models such as Llama-3.3-70B-Instruct  \cite{grattafiori2024llama} and Qwen-3 32B \cite{yang2025qwen3}.

\subsection{Question Generation}
\label{sec:prompt-qgen}

The system prompt shared across question generation strategies:

\begin{lstlisting}[style=prompt]
You are generating realistic financial customer
questions. Customers ask questions to a financial
assistant chatbot. Your questions must:
- Sound like a real customer typed them into a chat
  or search box
- Be answerable from the provided financial text
- Match the specified style and constraints EXACTLY

CRITICAL: Output ONLY the question text.
No explanations, no numbering, no quotes.
\end{lstlisting}

\noindent The user prompt conditions generation on attributes sampled from real query distributions (\S3.1):

\begin{lstlisting}[style=prompt]
Financial text:
---
[FINANCIAL TEXT SEGMENT]
---

Generate ONE question a [PERSONA] would ask about
the information above.

CONSTRAINTS:
- Question style: [STYLE_DESCRIPTION]
- Target length: approximately [WORD_COUNT] words
- Formality: [FORMALITY]

STYLE EXAMPLES for "[STYLE]":
[FEW-SHOT EXAMPLES]

Your question:
\end{lstlisting}

\noindent Dynamic fields are sampled per-question: \texttt{PERSONA} from \{retail customer, small business owner, financial advisor\}; \texttt{STYLE} from the distribution in Table~\ref{tab:qgen-key}; \texttt{WORD\_COUNT} from $\mathrm{LogNormal}(\mu{=}2.1, \sigma{=}0.55)$; \texttt{FORMALITY} from \{casual, neutral, formal\}. Two additional strategies produce 23\% of questions: contrastive pairs (two questions with different styles from the same passage, output as JSON) and template slot-filling.

\subsection{Single-Step Baseline}
\label{sec:prompt-singlestep}

The single-step ablation baseline generates both question and answer in one API call, without the multi-stage pipeline:

\begin{lstlisting}[style=prompt]
System: You are an expert financial Q&A generator.
Your task is to:
1. Read the numbered sources about financial topics
2. Generate a natural, standalone question based on
   one of the sources
3. Provide a well-cited answer using the source
   information

Important:
- The question should NOT refer to "the text",
  "the passage", or "the document"
- The question should be specific and answerable
  from the provided sources
- Include citations [1], [2], etc. when referencing
  source content
- If the sources don't contain enough information,
  respond with the unknown phrase
\end{lstlisting}

\begin{lstlisting}[style=prompt]
User: Given the following financial text sources,
generate a question-answer pair.

**Context Sources:**
[NUMBERED SOURCES]

**Instructions:**
1. Choose ONE source that contains interesting or
   important financial information
2. Generate a natural question about that source's
   content
3. Provide a concise answer citing the relevant
   sources with [N] notation
4. If sources don't support a good question,
   respond with: "I don't know."

**Output Format (JSON):**
{
    "question": "Your generated question here",
    "answer": "Your answer with [1], [2] citations"
}

Generate the question-answer pair:
\end{lstlisting}

\subsection{Answer Generation with Hint}
\label{sec:prompt-answergen}

The full pipeline generates answers with a hint identifying the gold source. This is the prompt used in Step~4 of the 5-step process (Sec. \ref{sec:sec-synth-qa}):

\begin{lstlisting}[style=prompt]
You are a helpful assistant that provides answers to
questions based on the provided sources. You will be
given a set of sources and a question. Your task is
to generate an answer that accurately reflects the
information in the sources, while also including
citations for any specific details referenced.

Example:
Context: The following numbered sources are provided.
---------------------
Source [1]: Our savings account offers 2.5% APY for
balances over $5,000.
Source [2]: Withdrawals from savings accounts are
limited to 6 per month before fees apply.
---------------------

Question: How can I earn 2.5% APY on my savings?
Answer: You can earn 2.5% APY by maintaining a
balance of over $5,000 in your savings account[1],
but remember that withdrawals are limited to 6 per
month before fees apply[2].

Instructions:
1. Answer the Question using only the information
   from the provided sources.
2. Include source citations using their
   corresponding numbers (e.g., [1]).
3. Every answer must contain at least one citation.
4. Only cite a source if you directly reference it.
5. Keep the answer concise and focused.
6. Use bulleted lists for clarity if multiple points
   are made.
7. If none of the sources are relevant, respond with
   "I don't know." and stop.

Context: The following numbered sources are provided.
---------------------
{{context|numbered}}
---------------------
Hint: The correct answer should be found in the
source number {{hint}}.

Question: {{question}}
Answer:
\end{lstlisting}

\noindent The \texttt{no\_hint} ablation variant is identical except the hint line is omitted, requiring the model to identify the correct source independently.

\section{Inference Prompt}
\label{sec:prompt-inference}

The following prompt is used at both training and inference time.
It uses Gemma~3's native chat tokens:

\begin{lstlisting}[style=prompt]
<start_of_turn>user
**Context:**
---------------------

Source [1]: ...
Source [2]: ...
...

---------------------

**Instructions:**
- Provide an answer based solely on the provided
  sources.
- Use only the provided context to construct your
  answer.
- Reference sources only when their information is
  explicitly used in your answer; include the
  corresponding source number as a citation
  (e.g., [1]).
- Every answer must include at least one source
  citation.
- If none of the provided sources are relevant,
  simply respond with 'I don't know.' and stop.

**Question:**

[QUESTION]

**Answer:**<end_of_turn>
<start_of_turn>model
\end{lstlisting}

\section{Evaluation Prompts}
\label{sec:prompts-eval}

\subsection{Citation Quality Judge (GPT-4.1)}

System message: \textit{``You are an expert evaluator of Citations in RAG system responses. Return ONLY valid JSON that strictly follows the requested format.''}

\begin{lstlisting}[style=prompt]
<rules>
- Evaluate the response solely based on the
  provided sources.
- Focus on answer quality and the effective use
  of citations.
</rules>
<format>
Return ONLY a JSON object that strictly adheres
to the schema below.
</format>
<structure>
---
QUESTION:
{question}

SYSTEM RESPONSE:
{response}

AVAILABLE SOURCES:
{sources}

1. Provide a brief evaluation summary.
2. Score the response (0-10) for:
- Source Relevance: How well the sources support
  the answer.
- Answer Quality: Clarity, correctness, and
  conciseness.
- Citation Usage: Appropriateness and accuracy of
  citations.
- Information Synthesis: How well the information
  is integrated.
- Faithfulness: Accuracy relative to the provided
  sources.
3. List key strengths, weaknesses, and improvement
   suggestions.
4. Provide an overall rating (0-10).
</structure>
Return ONLY a JSON object:
{ "scores": {...}, "analysis": {...},
  "overall_rating": <0-10> }
\end{lstlisting}

\noindent The five sub-scores are averaged to produce the Citation Quality composite (0--100) reported in Tables~\ref{tab:main} and~\ref{tab:curriculum-ablation}.

\subsection{Answer Quality Judge (JudgeLM 7B)}

JudgeLM~\cite{zhu2023judgelm} uses a comparative evaluation format, scoring the model response (Assistant~2) against the gold reference (Assistant~1):

\begin{lstlisting}[style=prompt]
You are a helpful and precise assistant for
checking the quality of the answer.

[Question]
{question}

[The Start of Assistant 1's Answer]
{reference}
[The End of Assistant 1's Answer]

[The Start of Assistant 2's Answer]
{single answer}
[The End of Assistant 2's Answer]

[System]
We would like to request your feedback on the performance of two AI assistants in response to theuser question displayed above. Please rate the helpfulness, relevance, accuracy, level of details of their responses. Each assistantreceives an overall score on a scale of 1 to 10, where a higher score indicates better overallperformance.Please first output a single line containing only two values indicating the scores for Assistant 1 and2, respectively. The two scores are separated by a space. In the subsequent line, please provide acomprehensive explanation of your evaluation, avoiding any potential bias and ensuring that theorder in which the responses were presented does not affect your judgment.

[Response]
10 
\end{lstlisting}

\end{document}